\documentclass[letterpaper, 10 pt, conference]{ieeeconf}  

\IEEEoverridecommandlockouts                              

\usepackage{graphics} 
\usepackage{epsfig} 
\usepackage{amsmath} 
\usepackage{amssymb}  
\usepackage{bm}
\usepackage{flushend}
\usepackage{soul}
\usepackage{dsfont}
\usepackage{amssymb}
\usepackage{float}
\usepackage[font=small]{caption}

\usepackage{hyperref}

\newcommand{\argmin}{\mathop{\mathrm{argmin}}}
 
\newcommand{\nbf}[1]{{\textbf{#1.}}}
\usepackage[noadjust]{cite}
\usepackage[font={footnotesize}]{caption}

\title{\LARGE \bf
AMEND: A Mixture of Experts Framework for Long-tailed Trajectory Prediction
}
\author{Ray Coden Mercurius$^{1,3}$, Ehsan Ahmadi$^{2,3}$, Soheil Mohamad Alizadeh Shabestary$^{3}$, Amir Rasouli$^{3}$
\thanks{$^{1}$ University of Toronto. Work done while at Huawei. {\tt\small ray.mercurius@mail.utoronto.ca}}%
\thanks{$^{2}$  University of Alberta, {\tt\small eahmadi@ualberta.ca}}
\thanks{$^{3}$ Noah's Ark Laboratory, Huawei, Canada. {\tt\small first.last@huawei.com}}}

\begin{document}

\maketitle
\begin{abstract}
Accurate prediction of pedestrians' future motions is critical for intelligent driving systems. Developing models for this task requires rich datasets containing diverse sets of samples. However, the existing naturalistic trajectory prediction datasets are generally imbalanced in favor of simpler samples and lack challenging scenarios. Such a long-tail effect causes prediction models to underperform on the tail portion of the data distribution containing safety-critical scenarios. Previous methods tackle the long-tail problem using methods such as contrastive learning and class-conditioned hypernetworks. These approaches, however,  are not modular and cannot be applied to many machine learning architectures. In this work, we propose a modular model-agnostic framework for trajectory prediction that leverages a specialized mixture of experts. In our approach, each expert is trained with a specialized skill with respect to a particular part of the data. To produce predictions, we utilise a router network that selects the best expert by generating relative confidence scores. We conduct experimentation on common pedestrian trajectory prediction datasets and show that our method improves performance on long-tail scenarios. We further conduct ablation studies to highlight the contribution of different proposed components.
\end{abstract}

\section{Introduction}
Trajectory prediction is a safety-critical task where the goal is to predict the future trajectories of the agents given their history information and the state of their surrounding environment. Relying on such information, the existing trajectory prediction models  \cite{yuan2021agentformer, Pourkeshavarz_2023_ICCV, Zhu_2023_CVPR, karim2023destine, amirloo2022latentformer} achieve promising performance on the benchmarks. However, they are suffering from low accuracy performance on long-tail challenging scenarios. 

As a result of the long-tail phenomenon, prediction models focus on more frequent (often simpler) scenarios and tend to put less emphasis on rarer challenging cases \cite{CFTP}. This limits the applicability of the existing approaches to practical intelligent driving systems. 

A commonly adopted approach to the long-tail problem in trajectory prediction is employing contrastive learning which aims to better organize latent features for more balanced training \cite{CFTP, FEND}. However, this scheme is not compatible with many existing architectures \cite{MTR, MGTR, Rasouli_2021_ICCV, Shi_2023_ICCV}, as they employ multiple encoded vectors in their latent space bottleneck. For example, one for each agent or road element. Moreover, contrastive learning can impose additional computational burden which is not desirable for practical systems. 

Alternatively, the long-tail problem can be addressed using multiple specialized experts, each focusing on a particular sub-task \cite{SADE}. This allows the model to equally pay attention to each subset of the data regardless of their distribution. A shortcoming of this solution, however, is the way the experts are aggregated which adds to computational overhead. 

To this end, we propose a novel framework \textbf{AMEND}: \textbf{A} \textbf{M}ixture of \textbf{E}xperts Framework for Lo\textbf{n}g-taile\textbf{d} Trajectory Prediction. Our framework is based on the divide-and-conquer technique, where a complex task can be decomposed into a set of simpler sub-tasks corresponding to sub-domains in the input space. For example, motion behaviour at intersections is very different from that on straight roadways.

Our approach follows a two-step training regiment. In the first phase,  we cluster the data into distinct sections with shared characteristics and then train an expert model on each cluster. In the next phase, we train a router network by ranking the performance of experts on the training data. During inference time, the router network scores each expert given the test sample, and based on the score, a selection module chooses which expert to use to generate predictions.

By directing the input to a single expert at a time, our approach avoids any additional computational cost during inference. In addition, our method is model-agnostic and modular, meaning that it treats the backbone model as black-box and only controls its inputs.

Our \textbf{contributions} are as follows:  We propose a novel mixture of experts framework for trajectory prediction. Our framework encourages the diversity of expert skill sets to mitigate the long-tailed distribution problem, while simultaneously avoids additional computational cost. We conduct empirical evaluations on common pedestrian trajectory benchmark datasets and highlight the advantage of our multi-expert method on predicting challenging scenarios. At the end, we perform ablation studies, showing the benefits of proposed modules on the overall performance.

\section{Related Work}
\subsection{Trajectory Prediction}
Trajectory prediction models aim to forecast the future positions of agents given their past trajectories and their surrounding context. There is a large body of literature in this domain, many of which are catered to pedestrian trajectory prediction \cite{Shi_2023_ICCV, Rasouli_2021_ICCV, rasouli_2023_icra, yuan2021agentformer, rasouli2023novel, Traj++}. These models rely on variety of architectures, such as recurrent networks \cite{Su_2021_IROS, Rasouli_2021_ICCV, Dendorfer_2021_ICCV}, graph neural networks \cite{Hasan_2022_ICRA, Shi_2021_CVPR, Traj++}, and transformers \cite{Shi_2023_ICCV, rasouli2023novel,rasouli_2023_icra, yuan2021agentformer} to effectively capture the complex contextual information. In this work we use Trajectron++ EWTA \cite{CFTP} as our baseline, which is a variation of \cite{Traj++}, a graph-based model. 

\subsection{Long-Tailed Learning}
Long-tailed learning seeks to improve the performance on tailed samples in imbalanced datasets and it is well studied in the computer vision domain \cite{LT_survey}. The re-balancing methods either oversample or undersample imbalanced classes, reweigh the loss function during training, or directly adjust the classification logits during inference to encourage the model to predict low-frequency classes \cite{rebalancing, bias_loss, bias_logits}. A shortcoming of re-balancing methods is that they only perform sample removal or duplication, without adding any new information. This issue is resolved in information augmentation methods, which create new training examples in the tail classes \cite{AV_scene_generation}. These methods, however,  work best with low-dimensional and simple data distributions, which are not the case for autonomous driving data \cite{AV_scene_generation}.

The long-tail problem has also been investigated in trajectory prediction. The authors of \cite{CFTP} utilise contrastive learning to separate the difficult scenarios from the easy ones in the latent space  allowing the model to better recognize and share information between difficult scenarios. FEND \cite{FEND} improves the contrastive learning framework by introducing artificial classes formed by clustering the encoded feature vectors of an autoencoder network. A shortcoming of these techniques is that they work with a single latent vector that captures all the scene information. State-of-the-art trajectory prediction architectures \cite{MTR, MGTR} employ multiple latent vectors at the bottleneck, such as one for each scene object, and therefore there is no singular feature vector to reshape according to the sample's class. 

FEND used a class-conditioned hypernetwork \cite{ha2017hypernetworks} decoder that allows dynamic and specialized decoder weights for different scenario types \cite{FEND}. However hypernetworks have many limitations, such as challenges in parameter initialization and complex architectures that must follow. In this work we propose a model-agnostic framework that relies on multiple experts in a computationally efficient fashion without the need for constrastive learning.

\subsection{Mixture of Experts}
Mixture of Experts (MoE) is a machine learning technique that utilizes several base learners, each one specialized on a particular sub-task \cite{adaptive_moe}. MoE differs from ensembling in that only one or a few experts are run for each input value and it can be restricted to only a portion of the model's architecture \cite{MOE_survey}. MoE is very effective in increasing accuracy without proportional increase in the computational cost. 

MoE has been applied to various sequence analysis tasks, such as natural language processing \cite{adaptive_moe, outrageously_lnn}. Of interest,  the approach proposed in \cite{zhou2022mixtureofexperts} uses a novel routing algorithms,  where instead of each input being routed to the top-$k$ experts, each expert selects its top-$k$ inputs. In our work, we adopt a routing network  that is trained to score the experts based on the input sample, and in turn uses the best expert to generate trajectory output.

\section{Problem Formulation}
Given input information consisting of trajectory histories of $N$ agents in the scene $x_{t}^{i} \in \mathbb{R}^{2}$, $i \in [1,N], t \in [1-T_{hist},0]$, where $x_{t}^{i}$ is the 2D coordinates of agent $i$ at timestep $t$ and $T_{hist}$ is the number of history timesteps, our task is to predict the agents' future trajectories  $y_{t} \in \mathbb{R}^{2}$, $t \in [1,T_{pred}]$, where  $y_{t}$ is the coordinates of the agents at time $t$ and $T_{pred}$ is the number of prediction timesteps.

\begin{figure*}[t]
    \centering
    \includegraphics[width=1\textwidth]{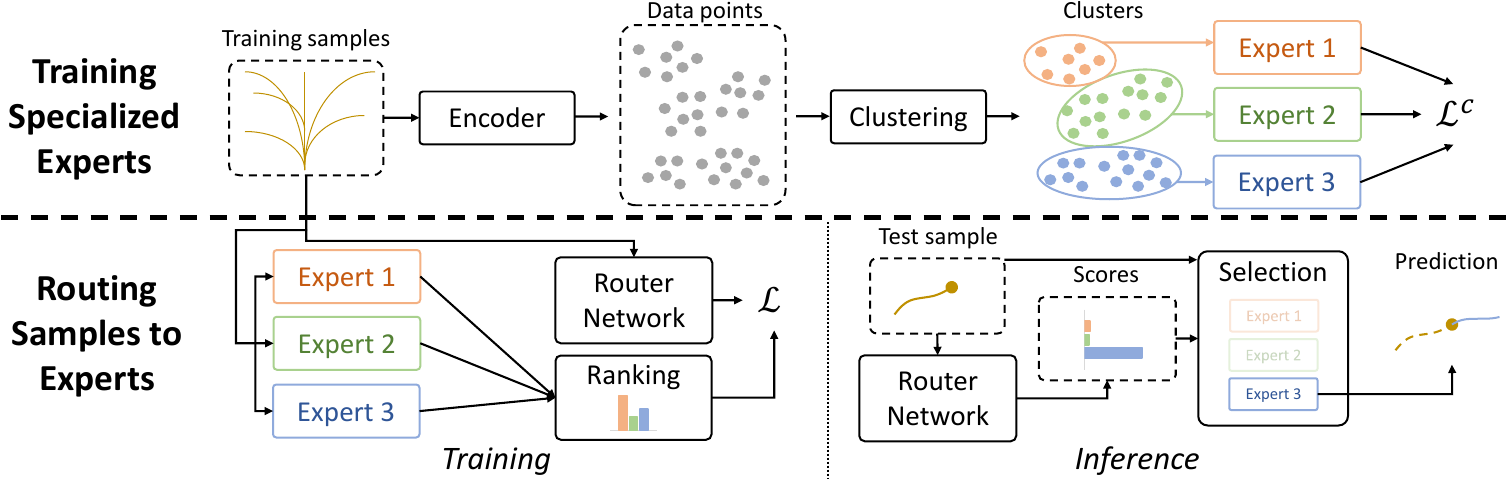}
    \caption{Overview of the proposed approach. We cluster the data samples based on the latent vector of an encoder network. During training of the experts the loss function is adjusted so that each expert focuses on a particular sample cluster. Next, we calculate the relative performance rankings of the experts, which are used to generate targets to train the router network. At inference we use the router network to select best expert to generate the predictions.}
    \label{fig:model}
\end{figure*}

\section{Methodology}
In this section, we describe our proposed solution to address the long-tailed learning problem in trajectory prediction. An overview of the proposed method is shown in \autoref{fig:model}. Individual parts of the method are described below.

\subsection{Training Specialized Experts}
\label{method_training_epxerts}
Our objective is to train multiple experts on a diverse dataset exhibiting a long-tailed distribution, assigning each expert to concentrate on distinct data patterns. By segmenting the learning task into simpler uniform sub-tasks, we facilitate more effective learning for each expert. Due to the lack of explicit labels, we will employ unsupervised learning strategies to segregate the dataset into sub-tasks.

We divide the original dataset $D$ into $C$ mutually exclusive subsets called $D_{1:C}$. We assign a unique expert to each subset to create $C$ experts represented by $E_{1:C}$. The samples within each subset should be adequately similar to allow expert specialisation.

\nbf{Clustering in latent space} Processing and clustering high dimensional inputs is challenging, hence, we use an encoder network and cluster the dataset based on the feature encodings. We select our encoder to be that of Trajectron++ EWTA \cite{CFTP} trained on the same dataset. We chose the encoder of a trajectory prediction model as it naturally embeds input information for a purpose that aligns with our final end-goal of forecasting trajectories.

Effective encoders only keep information relevant for the training task and map similar samples nearer in the latent space. Therefore, our latent space clusters should contain scenarios similar from a trajectory prediction standpoint. To achieve this, we perform K-means clustering on the latent vector of the encoder. Our approach differs from previous sample clustering methods for trajectory prediction, such as FEND \cite{FEND}, which forms clusters on the latent space of an autoencoder applied to individual trajectories. We include additional contextual information in our latent space, such as nearby agent behaviour.

\nbf{Loss function} Our goal is to force the model during training to focus more on specific subdomains of data containing unique prediction patterns, but without losing  generalization. To satisfy this trade-off, we train experts on all samples using a modified loss function that assigns more weight to samples belonging to the expert's assigned cluster. The training loss function of an expert over a batch is defined as:
\begin{equation}
\mathcal{L}^c=\frac{1}{B}\sum_{i=1}^{B}(\mathbf{\mathds{1}}[x_{i}\in D_{c}](1+\alpha) + \mathbf{\mathds{1}}[x_{i}\not\in D_{c}](1-\alpha))\mathcal{L}^{c}_{i},
\end{equation}
where $B$ is the batch size, $x_{i}$ is a training example, $c$ denotes a particular expert, $\mathcal{L}^{c}_{i}$ is the original loss of expert $E_{c}$ on sample $x_i$ in our baseline model, $\mathbf{\mathds{1}}(R)$ is the identity function that returns 1 when the condition $R$ is satisfied and otherwise 0, and $\alpha$ is a hyperparameter. Setting $\alpha=1$ results in mutually exclusive input data subsets for the experts, while setting $\alpha=0$ results in identical training.

\subsection{Routing Samples to Experts}
During inference, our challenge is how to assign the test samples to the best expert to perform the forward pass and generate predictions. We propose to use a router network to predict the aptitude of each expert on a given test sample. The router network takes in the input information and outputs a confidence score $p_{1:C}$ for each of the $C$ experts, where $\sum_{c=1}^{C}p_{c}=1$, indicating the probability of that expert being the best expert for the given sample. Then, the expert with the highest confidence score is selected. The architecture of the router network consists of an encoder network (adopted from the baseline) followed by two fully connected layers. 

For router training, to identify the best-performing expert for a given sample, we rely on Average-Displacement-Error (ADE) and Final-Displacement-Error (FDE) metrics,
\begin{equation}
c_{best}=\argmin_{c}\text{R}_{FDE}^{c}+\text{R}_{ADE}^{c},
\end{equation}
where $c_{best}$ indicates the best expert that has the lowest combined FDE and ADE among all experts, and $R^{c}_{FDE}, R^{c}_{ADE} \in \mathbb{N}$ are the rankings of the expert $E_{c}$ on the $FDE$ and $ADE$ metrics, respectively,  with $R=1$ indicating the best performing expert. We use cross-entropy loss to train the router network, with the target being a one-hot vector where index $c_{best}$ is one.

At inference step,  we direct the inputs to the expert with the highest confidence score in a winner-takes-all aggregation scheme. An advantage of this is that the forward pass is only computed for a single expert, hence, the inference computational cost does not scale with the number of experts.

\begin{table*}
\caption{Quantitative evaluation on long-tail scenarios for the ETH-UCY benchmark computed based on the weighted average of a five-fold evaluation. Results are reported on $\text{minADE}_{20}$/$\text{minFDE}_{20}$. Top $\alpha\%$ split refers to performance on the highest percentile of challenging scenarios. $\text{VaR}_{97}$  refers to the $0.97$ quantiles of the error distribution. Relative metrics are provided in the last three columns as a normalized measure of the model performance on the long-tail. \textbf{Bold} numbers indicate the best performance for each metric. * indicates the model without publicly available code, hence, it is not considered in ranking. For all metrics lower value is better.}
\vspace{-0.4cm}
\label{tab:long-tail}
\begin{center}
\resizebox{\textwidth}{!}{%
\begin{tabular}{c||c|c|c|c|c|c|c|c}
\rule[-3ex]{0pt}{0ex}
  & \multicolumn{5}{c|}{$\text{minADE}_{20}$/$\text{minFDE}_{20}$  }  & \multicolumn{3}{c}{ $\frac{\raisebox{.03in}{$\text{minADE}_{20}$}}{\raisebox{-.05in}{$\text{minADE}_{20}^{All}$}}$/$\frac{\raisebox{.03in}{$\text{minFDE}_{20}$}}{\raisebox{-.05in}{$\text{minFDE}_{20}^{All}$}}$} \\
 \cline{2-9}
 \rule{0pt}{2ex}
Method & $\text{Top 1 \%}$ & $\text{Top 3 \%}$ & $\text{Top 5 \%}$  &  $\text{All}$ & $\text{VaR}_{97}$ & Top $\text{1\%}$ & Top $\text{3\%}$ &   Top $\text{5\%}$\\

\hline
\rule{0pt}{2ex}
Traj++ & 0.58/1.23 & 0.65/1.42 & - & 0.21/0.41 & 0.78/1.97 & 2.8/3.0  & 3.5/3.1  & - \\
\rule{0pt}{2ex}
Traj++ EWTA & 0.45/0.89 & \textbf{0.47}/0.99 & \textbf{0.43}/0.89 & \textbf{0.18}/0.32  & \textbf{0.55}/1.18 & 2.5/2.8  & \textbf{2.6}/3.1  & \textbf{2.4}/2.8   \\
\rule{0pt}{2ex}
contrastive & 0.45/0.81 & 0.48/0.99 & 0.44/0.88 & \textbf{0.18}/0.32 & 0.56/1.18 & 2.5/2.5  & 2.7/3.1 & \textbf{2.4}/2.8  \\ \hline
\rule{0pt}{2ex}
FEND*  & 0.38/0.74 & 0.40/0.85 & 0.37/0.76 & 0.17/0.32 & - & 2.2/2.3 & 2.4/2.7  & 2.2/2.4 \\
\hline\hline
\rule{0pt}{2ex}
\textbf{AMEND}  & \textbf{0.43}/\textbf{0.75} & 0.48/\textbf{0.91} & \textbf{0.43}/\textbf{0.80} & \textbf{0.18}/\textbf{0.31} & \textbf{0.55}/\textbf{1.11} & \textbf{2.4}/\textbf{2.4} & 2.7/\textbf{2.9} & \textbf{2.4}/\textbf{2.6}  \\
\end{tabular}
}
\end{center}
\vspace{-0.4cm}
\end{table*}


\section{Experiments}

\subsection{Experimental Setup}
\nbf{Datasets}
We evaluate the models on ETH-UCY, which are bird's-eye-view pedestrian benchmark datasets \cite{eth, ucy}. The datasets contain challenging scenarios, such as crowded scenes with complex agent-to-agent interactions. Following the previous works \cite{Shi_2023_ICCV, Traj++, yuan2021agentformer}, we average our final results over a five-fold cross-validation scheme, with four splits utilized for training and one for test. For the ETH-UCY datasets, $T_{pred}$ is 12, $T_{hist}$ is 8 and the samples are collected at $2.5Hz$ ($\Delta t = 0.4 s$).

\nbf{Metrics}
We use the common performance metrics for trajectory prediction \cite{Shi_2023_ICCV, Traj++, yuan2021agentformer}, namely Average-Displacement-Error (ADE) and Final-Displacement-Error (FDE), and report the minimum error across $K=20$ predictions. 

For evaluating performance on tail samples, we use the following methods:

\noindent
\textit{i) Scenario Difficulty Ranking}: We evaluate the model's performance on the top 1$\%$ and 5$\%$ difficult scenarios. To rank the scenarios by difficulty we utilise the errors of a simple Kalman filter \cite{CFTP}.

A drawback of this metric is that the definition of the tailed scenarios is dependant on the model used to judge difficulty. Therefore, different models might underperform on different scenarios and the definition of the data tail might vary \cite{FEND}. Furthermore, simply measuring errors on the set of challenging scenarios does not properly capture the changes in the distribution of errors across the dataset. It is possible for a trade-off to occur in which the model's performance deteriorates on other scenarios.
\noindent
\textit{ii) Error Distribution Quantiles}: We use an alternative  metric which directly measures the magnitude of the tail of the error distribution. This is the error on the worst performing samples according to the model. We adopt the value-at-risk (VaR) metric. VaR$_{\alpha}$ refers to the $\alpha^{th}$ quantile of the error distribution, where $\alpha \in (0, 1)$:
\begin{equation}
\text{VaR}_{\alpha}(E) = \text{inf} \{ e \in E: P(E \geq e) \leq 1-\alpha \}.
\end{equation}
\noindent
More formally,  it is the smallest error $e$ such that the probability of observing error larger than $e$ is smaller than $1$-$\alpha$, where $E$ is the distribution of errors. We measure VaR at $0.97$.

\nbf{Models}
We report the results on two baseline models, Trajectron++ (Traj++) \cite{Traj++} and its variation Trajectron++ EWTA (Traj++ EWTA for short) \cite{CFTP}, which replaces Conditional Variational Auto Encoder (CVAE) module in Traj++ with the multi-hypothesis networks trained with Evolving Winner-Takes-All (EWTA). A variation of Traj++ EWTA with contrastive loss, denoted at contrastive, is also reported. Moreover, we report on state-of-the-art model FEND \cite{FEND}. However, since this model does not have a publicly available code, we do not consider it in the ranking of the models.

\nbf{Implementation Details}
Our main model and the router network follow the same training schedule. We train our main model for 300 epochs, with the EWTA schedule starting at $\textup{K}=20$,  decaying self-adaptively by 0.8 if accuracy metrics do not improve, and ending at $\textup{K}=1$. The router softmax temperature is set to 1.  The dimension of our router decoder's hidden layer is 232. All other training details are kept the same as in the baseline model \cite{CFTP}. 

\nbf{Data Preprocessing}
To eliminate the impact of arbitrary scale, we normalize the dimensions similar to \cite{CFTP}. The trajectory coordinates are divided by their standard deviation in the training dataset. Additionally, we  normalize the headings by rotating the inputs so that the last known direction of the agent points in the positive y-axis. The opposite rotation is then applied to the model's outputs. With this orientation normalization the trajectory end-points capture the general intention of the agents.

\subsection{Experimental Variations}

\nbf{Clustering on trajectory endpoints} We experiment with modifying the clustering algorithm used to partition the dataset. Instead of clustering on the latent space, we perform K-Means \cite{macqueen1967kmeans} clustering on endpoints of the ego-vehicle's future trajectories. We denote the model that uses this clustering as Trajectory. Empirically, we find that this results in partitions based on modality, such as turn type and velocity. A shortcoming is that it only utilises information from the ego's trajectory, and ignores other scene information such as interactions with nearby agents or potential map info.

\nbf{Cluster Assignment} In this experiment we replace the router network with a heuristic algorithm to generate expert confidence scores. We rely on the principle that an expert should perform best on examples most similar to its assigned training cluster. We utilize the distance between the embedding of two arbitrary samples in the latent space as a proxy for how similar the samples are. Therefore, given a sample, we calculate the latent distance between itself and the cluster centroid of each expert as a rough approximation to the confidence of each expert on that sample. Recall that each expert $E_{1:C}$ focused on a unique subset of data $D_{1:C}$ during training, and is associated with a unique cluster centroid denoted by $\phi_{1:C}$. Confidence scores are generated via a Softmax operation on these distances as follows:

\begin{equation}
p^{c}_{i} = \frac{\text{exp} (\text{-dist}(\rho(x_{i}), \phi_{c}))}{\sum_{j=1}^{C} \text{exp} (\text{-dist}(\rho(x_{i}), \phi_{j}))},
\end{equation}

\noindent
where ${i}$ denotes sample index, ${c}$ denotes the expert index, $\rho(\cdot)$ is our encoder network and $\phi_{j}$ is a cluster centroid. We denote this model as Cluster-based.

\subsection{Long-tailed Prediction}
We compare our method to trajectory prediction models catered to long-tailed prediction. All models utilize variations of Trajectron++ \cite{Traj++} as their backbone model which provides a fair comparison. As shown in \autoref{tab:long-tail}, compared to contrastive, on Top \% metrics,  our framework achieves better performance on most cases, improving minADE and minFDE by up to 9$\%$. Highest minADE improvement is achieved on Top 1\% consisting of most challenging scenarios while highest minFDE improvement is on top 5\%. 

From the relative metrics (the last two coloumns), we can see that given our higher performance ratio on challenging scenarios, compared to the average case,  our model achieves a more balanced performance across different difficulty levels. This indicates  that our approach, which consists of allocating more training resources to specific prediction patterns via specialized modules is especially helpful for complex patterns. We verify the soundness of our improvements by comparing the VaR metric which measures the error distribution quantiles. Achieving lower values on VaR metrics overall means that the largest prediction errors given by our model across the dataset is smaller than the largest errors of other models. For the proposed model, the improvement is apparent for final error (by 6\%). This indicates that overall, our model was successful by preventing the error migration from one domain to another. 

\subsection{Training Diverse Experts}
\begin{figure}[tp]
    \centering 
\includegraphics[width=\columnwidth]{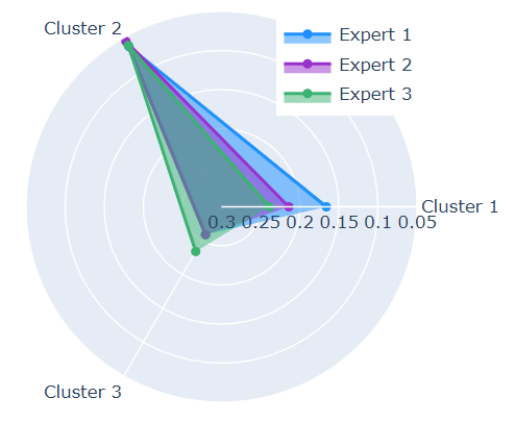}
    \caption{Radar plot showing the FDE of each expert on the different cluster splits of the ETH-UCY Hotel test dataset showing significant variations in the performance of the experts. The best performing expert for each cluster tends to be the one that was assigned to it during training.}
    \label{fig:radar}
\end{figure}

In \autoref{fig:radar}, we show the performance of each expert on test samples across different clusters. As expected, the experts perform best in the cluster of scenarios assigned to them during training. Note that the best performing expert for each cluster tends to be the expert that was assigned to that cluster.

In \autoref{tab:cls_basis_routing_method}, we compare different training methods to create specialized experts. For clustering (on top) we compare Trajectory, which is a model that clusters scenarios based on the final endpoints of the ego-trajectory to ours (Latent Feature). Here, we can see that our approach outperforms Trajectory, especially on $\text{VaR}$. Such improvement can be due to added information captured in the latent space,  accounting for factors,  such as interactions between the agents.

\begin{table}[t]
\caption{The effect of clustering basis (top rows) and the routing method (bottom rows) on the overall performance and the long-tailed performance of the AMEND model. The $\text{minADE}_{20}$/$\text{minFDE}_{20}$ are reported based on the average of 5-fold evaluation for the ETH/UCY dataset. The * indicates the default setting.}\vspace{-0.4cm}
\label{tab:cls_basis_routing_method}
\begin{center}
\resizebox{1\columnwidth}{!}{%
\begin{tabular}{c||c|c|c|c|c}
\  & $\text{Top 1 \%}$ & $\text{Top 3 \%}$ & $\text{Top 5 \%}$ & $\text{All}$ & $\text{VaR}_{97}$\\
\hline
\hline
\rule{0pt}{2ex}
$\text{Trajectory}$  & 0.45/0.81 & 0.50/0.99 & 0.44/0.87 & 0.18/0.31 & 0.57/1.21 \\
\rule{0pt}{2ex}
$\text{Latent Feature*}$  & 0.43/0.75 & 0.48/0.91 & 0.43/0.80 & 0.18/0.31 & 0.55/1.11 \\
\hline
\rule{0pt}{2ex}
Cluster-based  & 0.42/0.85 & 0.47/0.94 & 0.42/0.82 & 0.18/0.30 & 0.57/1.13 \\
\rule{0pt}{2ex}
Router Network*  & 0.43/0.75 & 0.48/0.91 & 0.43/0.80 & 0.18/0.31 & 0.55/1.11 \\
\end{tabular}
}
\end{center}\vspace{-0.5cm}
\end{table}
\begin{table}[t]
\caption{The accuracy of routing method in selecting the expert that performs best on a test sample. Best performing expert is defined as the one that achieves the lowest error metrics (ADE/FDE). Errors per sample are averaged over sixteen trials to reduce uncertainty. The * indicates our default approach. Higher value means better.}\vspace{-0.4cm}
\label{tab:routing_accuracy}
\begin{center}
\resizebox{0.9\columnwidth}{!}{%
\begin{tabular}{c||c|c|c}
\ & Random & Cluster-based & Router Network*\\
\hline
\rule{0pt}{2ex}
ETH & - & 0.37/0.38  & 0.34/0.34 \\
Hotel & - & 0.40/0.36 & 0.40/0.39  \\
Univ & - & 0.30/0.30 & 0.36/0.35 \\
Zara1 & - & 0.34/0.34 & 0.34/0.35 \\
Zara2 & - & 0.50/0.43 & 0.49/0.44 \\
\hline
\rule{0pt}{2ex}
Avg & 0.33/0.33 & 0.38/0.36 & 0.39/0.37 \\

\end{tabular}
}
\end{center}\vspace{-0.5cm}
\end{table}




\subsection{Expert Selection}
In \autoref{tab:cls_basis_routing_method} (bottom) we compare our main model which selects experts with the router network (Router Network), to an alternative approach which selects the expert who's sample cluster is predicted to contain the test sample (Cluster-based). Here, we can see that  utilising the router network generally generates better results.

We further investigate the discrepancy between the performance of our routing approach compared to the clustering technique. For this experiment, 
we report the results in terms of accuracy of the methods for predicting which expert would perform best. The results are summarized in \autoref{tab:routing_accuracy}. Here, the baseline for comparison is random routing, which has a $33\%$ chance of being correct since we have 3 experts. The strong performance of Cluster-based relative to random routing supports the hypothesis that the best performing expert for samples within a particular cluster is usually the expert specialized for the cluster. However, its lower performance compared to our Router-Network approach, suggests that there are exceptions to this rule, which our router has successfully learned. This supports the idea of using neural networks to map the complex distribution of relative expert strength across the data space for routing.

\section{Conclusion}
In this paper, we tackled the long-tail pedestrian prediction problem by formulating a Mixture of Experts framework. We proposed a novel two-stage training scheme in which we first train specialized experts on sub-tasks within the data, and second use the experts to train a routing network for scoring the experts at inference time. We demonstrated that clustering the dataset and focusing each expert's resources on a partition creates specialized skills which can be utilised to generate accurate predictions. We conducted extensive experimental evaluation on common pedestrian trajectory benchmark datasets, outperforming the  previous methods on challenging tailed samples. We further highlighted the effectiveness of our proposed modules via ablation studies.

\bibliographystyle{./IEEEtran.bst}
\bibliography{./IEEEabrv,./ref}

\end{document}